\documentclass[A4,11pt]{article}

\usepackage{times}
\usepackage{latexsym}

\usepackage{pstricks}
\usepackage{amssymb}
\usepackage{verbatim}
\usepackage{fullname}

\newcommand{\spc}[1]{\rule{#1}{0mm}}
\newcommand{\complexity}[0]{{\mathcal O}}	
\newcommand{\aTuple}[1]{\langle{#1}\rangle}	
\newcommand{\aSet}[1]{\{{#1}\}}			
\newcommand{\tapnum}[1]{^{(#1)}}	
\newcommand{\vspc}[1]{\rule{0mm}{#1}}
\def\eps{\varepsilon}			
\newcommand{\JOIN}[1]{\Join_{\aSet{#1}}}		

\newcommand{\mtchltr}[1]{{\large\bf #1}}
\newcommand{\TextExample}[1]{\vspace{1ex}\begin{minipage}[t]{120mm}{\sf\small #1}\end{minipage}\\}

\newsavebox{\BoxNUS}
\savebox{\BoxNUS}{{\small\tt[\spc{-0.4ex}\^\_\spc{-0.2ex}]}}
\def\NUS{\usebox{\BoxNUS}}

\begin{document}


\title{\vspace{-15mm}
  Acronym-Meaning Extraction from Corpora \\
  Using Multitape Weighted Finite-State Machines \vspace{1.5ex} \\
  {\Large ---~ Research Report ~2006 / 019 ~---}
}

\author{
  Andr\'e Kempe   \vspace{2ex} \\
  Xerox Research Centre Europe ~--~ Grenoble Laboratory \\
  6 chemin de Maupertuis ~--~ 38240 Meylan ~--~ France \\
  {\small\sf andre.kempe@xrce.xerox.com} ~--~ {\small\sf http://www.xrce.xerox.com}
}

\date{
  July 26, 2006
}

\maketitle

\thispagestyle{empty}


\begin{abstract}

The automatic extraction of acronyms and their meaning from corpora
is an important sub-task of text mining.
It can be seen as a special case of string alignment,
where a text chunk is aligned with an acronym.
Alternative alignments have different cost,
and ideally the least costly one should give the correct meaning of the acronym.

We show how this approach can be implemented
by means of a 3-tape weighted finite-state machine (3-WFSM)
which reads a text chunk on tape 1 and an acronym on tape 2,
and generates all alternative alignments on tape 3.
The 3-WFSM can be automatically generated from a simple regular expression.
No additional algorithms are required at any stage.
Our 3-WFSM has a size of 27 states and 64 transitions,
and finds the best analysis of an acronym in a few milliseconds.

\end{abstract}


\section{Introduction
  \label{sec:intro}}

The automatic extraction of acronyms and their meaning from corpora
is an important sub-task of text mining.
We will refer to it by the term {\it acronym-meaning extraction}\/.

Much work has been done on it. To mention just some:
%
\namecite{yeates+al:2000}
matched an acronym against a text chunk,
thus producing different candidate definitions for the acronym.
They alternatively tried heuristic approaches, naive Bayes learning,
and compression (i.e., shortest description length) to select the best candidate.
\namecite{pustejovsky+al:2001}
added shallow parsing, and matched an acronym against a parsed (i.e., structured) text chunk.
\namecite{schwartz+hearst:2003}
used a heuristic algorithm to deterministically find an acronym's meaning.

The task can bes seen as a special case of string alignment
between a text chunk and an acronym.
For example, the text chunk ``{\sf\small they have many hidden Markov models}''
can be aligned with the acronym ``{\sf\small HMMs}'' in different ways,
such as
``{\small\sf they have many \mtchltr{h}idden \mtchltr{M}arkov \mtchltr{m}odel\mtchltr{s}}''
or
``{\small\sf t\mtchltr{h}ey have \mtchltr{m}any hidden \mtchltr{M}arkov model\mtchltr{s}}''.
Alternative alignments have different cost,
and ideally the least costly one should give the correct meaning of the acronym.
String alignment admits in general four different edit operations:
insertion, deletion, substitution, and preservation of a letter
\cite{wagner+fischer:1974,pirkola+al:2003}.
In the case of acronym-meaning extraction, only deletion and preservation can occur.
String alignment has a worst-case time complexity of $\complexity(|s_1||s_2|)$,
with $|s_1|$ and $|s_2|$ being the lengths of the two aligned strings, respectively.
This also holds for the present special case.

\smallskip

The purpose of this report is to show how the alignment-based approach to acronym-meaning extraction
can be implemented by means of a 3-tape weighted finite-state machine (3-WFSM).
The 3-WFSM will read a text chunk on tape 1 and an acronym on tape 2,
and generate all possible alignments on tape 3,
inserting dots to mark which letters are used in the acronym.
For the above example this would give ``{\small\sf they have many .hidden .Markov .model.s}'',
among others.

The 3-WFSM can be automatically generated from a clear and relatively simple regular expression
that defines the task in as much detail as we wish.
Both the generation and the application of the 3-WFSM are done by (more or less) standard operations
\cite{kempe+champarnaud+eisner:2004a},
that are available in finite-state tools such as WFSC
\cite{kempe+al:2003}.
No additional algorithms are required at any stage,
which reduces the development time to a few hours.


\section{Preprocessing the Corpus}

Using basic UNIX utilities,
we first extract from a corpus all sentences that contain acronyms in brackets,
such as

\TextExample{
  Between a hidden Markov model (HMM) and a weighted finite-state \\
  \spc{1ex} automaton (WFSA) there is a correspondence.}

\noindent
Then, we split these sentences into pairs consisting of an acronym and the text chunk
that precedes it
(starting from the sentence beginning or form the preceding acronym, respectively).
For the above example, this is

\TextExample{
  Between a hidden Markov model\spc{10ex}HMM \\
  and a weighted finite-state automaton\spc{6ex}WFSA}

\noindent
Next, we normalize these pairs:
capital letters are transformed into small ones,
and separators into underscores:

\TextExample{
  between\_a\_hidden\_markov\_model\spc{10ex}hmm \\
  and\_a\_weighted\_finite\_state\_automaton\spc{5.8ex}wfsa}


\section{Constructing an Acronym-Meaning Extractor}

We start by compiling a 4-WFSM over the real tropical semiring
$\aTuple{\mathbb{R}_{\geq 0}\cup\aSet{\infty}, \min, +, \infty, 0}$,
from the expression
\begin{eqnarray}
  A_1\tapnum{4} & \;=\; & \left(\;\;\vspc{2.8ex}
	\left\langle\vspc{2.1ex} \aTuple{\eps,\eps,\;{\bf .}\;,\eps}, 0 \right\rangle \;
			\left\langle\vspc{2.1ex} \aTuple{\NUS,\NUS,\NUS,{\tt a}}_{\aSet{1=2=3}} , 0 \right\rangle
	\;\;\;\cup\;\;\;
	\right.
	\nonumber \\
	& ~ &
	\left.\spc{10ex}
	\left\langle\vspc{2.1ex} \aTuple{\NUS,\eps,\NUS,{\tt i}}_{\aSet{1=3}} , 0 \right\rangle
	\;\;\;\cup\;\;\;
	\left\langle\vspc{2.1ex} \aTuple{\_,\eps,\_,\_} , 0 \right\rangle
	\vspc{2.8ex}\;\;\right)^{\large *}
\end{eqnarray}

\noindent
where $\NUS$ is a {\it symbol class} accepting any symbol except underscore,
$\eps$ represents the empty string,
$\aSet{1\!=\!2\!=\!3}$ a constraint
requiring the different $\NUS$ on tapes $1$ to $3$ to be instantiated by the same symbol
\cite{nicart+al:2006a},\footnote{
  Roughly following \cite{kempe+champarnaud+eisner:2004a},
  we employ here a simpler notation for constraints than in~\cite{nicart+al:2006a}.
}
and the $0$'s are weights.
We use a superscript $\tapnum{n}$ to indicate the arity of a $n$-WFSM
\cite{kempe+champarnaud+eisner:2004a}.

If we apply $A_1\tapnum{4}$ with tapes 1 and 2 to a normalized text chunk and a corresponding acronym,
respectively, and generate (neutrally-weighted) alternative analyses from tapes 3 and 4,
we obtain, for example

{\scriptsize
\begin{verbatim}
1,2> they_have_many_hidden_markov_models      hmms

3,4> t.hey_have_.many_hidden_.markov_model.s    iaii_iiii_aiii_iiiiii_aiiiii_iiiiia     0
3,4> t.hey_have_.many_hidden_markov_.model.s    iaii_iiii_aiii_iiiiii_iiiiii_aiiiia     0
3,4> t.hey_have_many_hidden_.markov_.model.s    iaii_iiii_iiii_iiiiii_aiiiii_aiiiia     0
3,4> they_.have_.many_hidden_.markov_model.s    iiii_aiii_aiii_iiiiii_aiiiii_iiiiia     0
3,4> they_.have_.many_hidden_markov_.model.s    iiii_aiii_aiii_iiiiii_iiiiii_aiiiia     0
3,4> they_.have_many_hidden_.markov_.model.s    iiii_aiii_iiii_iiiiii_aiiiii_aiiiia     0
3,4> they_have_many_.hidden_.markov_.model.s    iiii_iiii_iiii_aiiiii_aiiiii_aiiiia     0
\end{verbatim}
}

\noindent
On tape 3, letters of the text chunk which are used in the acronym, are preceded by a dot.
Tape 4 shows the performed operations:
{\tt a} meaning ``acronym letter'', {\tt i} meaning ``ignored letter''.
All analyses have neutral weight 0.

By means of XFST~\cite{karttunen+al:1998,beesley+karttunen:2003}
we generate from the following regular expression
a 2-FSM (i.e., a non-weighted transducer),
${A_2^\prime}\tapnum{2}$, that defines the operations more precisely:
\begin{equation}
\begin{minipage}{130mm}
{\scriptsize
\begin{verbatim}
regex
    [%_|a|u|i|g|G|1|2|3|4|5|6|7|8]*
.o. [ i -> u ||  a [ i | %_ ]* %_         _  i* [ .#. | %_ ] ]
.o. [ i -> g ||                           _  i* a  ]
.o. [ g -> G ||  [ .#. | %_ ]             _  ]
.o. [ a -> 1 ||  [ .#. | %_ ]             _  ]
.o. [ a -> 2 ||  [ .#. | %_ ] \%_         _  ]
.o. [ a -> 3 ||  [ .#. | %_ ] \%_^2       _  ]
.o. [ a -> 4 ||  [ .#. | %_ ] \%_^3       _  ]
.o. [ a -> 5 ||  [ .#. | %_ ] \%_^4       _  ]
.o. [ a -> 6 ||  [ .#. | %_ ] \%_^5       _  ]
.o. [ a -> 7 ||  [ .#. | %_ ] \%_^6       _  ]
.o. [ a -> 8 ||  [ .#. | %_ ] \%_^7 \%_*  _  ]
.o. [%_|a|u|i|g|G|1|2|3|4|5|6|7|8]*
\end{verbatim}
}
\end{minipage}
\end{equation}

\noindent
In this expression, all word-initial {\tt i} are replaced by {\tt u} (``unused word'')
if no letter of this word is used in the acronym,
but if letters of preceding words are used.
Then, all {\tt i} of a used word are replaced by {\tt g} (``gap letter'')
if they are followed by an {\tt a} (``acronym letter'') in the same word.
Next, word-initial {\tt g} are replaced by {\tt G} (``word-initial gap letter'').
Finally, all {\tt a} are replaced by a number {\tt 1} to {\tt 8},
expressing their position in the word.
Positions higher than 8 are marked as {\tt 8}.

Furthermore, we generate with XFST another 2-FSM, ${A_3^\prime}\tapnum{2}$,
that deletes all letters of leading unused words, and the adjacent underscores:
\begin{equation}
\begin{minipage}{130mm}
{\scriptsize
\begin{verbatim}
regex
    [%.|%_|a|b|c|d|e|f|g|h|i|j|k|l|m|n|o|p|q|r|s|t|u|v|w|x|y|z]*
.o. [ \%. -> 0 || .#. \%.* _ \%.* %_ ]
.o. [  %_ -> 0 || .#. _ ]
.o. [%.|%_|a|b|c|d|e|f|g|h|i|j|k|l|m|n|o|p|q|r|s|t|u|v|w|x|y|z]*
\end{verbatim}
}
\end{minipage}
\end{equation}

These two non-weighted 2-FSMs are transformed into 2-WFSMs,
$A_2\tapnum{2}$ and $A_3\tapnum{2}$,
with neutral weight,
and are joined
\cite{kempe+champarnaud+eisner:2004a}
with the previously compiled 4-WFSM $A_1\tapnum{4}$~:
\begin{equation}
  A_4\tapnum{6} =  \left( A_1\tapnum{4} \JOIN{3=1} A_3\tapnum{2} \right) \JOIN{4=1} A_2\tapnum{2}
\end{equation}

In the resulting $A_4\tapnum{6}$, we have a modified form of tape 3 on tape 5 (describing analyses),
and a modified form of tape 4 on tape 6 (describing operations).
If we apply tape 1 to a normalized text chunk and
tape 2 to a corresponding acronym,
we obtain from tapes 5 and 6 for example

{\scriptsize
\begin{verbatim}
1,2> they_have_many_hidden_markov_models     hmms

5,6> t.hey_have_many_hidden_.markov_.model.s   G2ii_uiii_uiii_uiiiii_1iiiii_1gggg6     0
5,6> t.hey_have_.many_hidden_markov_.model.s   G2ii_uiii_1iii_uiiiii_uiiiii_1gggg6     0
5,6> t.hey_have_.many_hidden_.markov_model.s   G2ii_uiii_1iii_uiiiii_1iiiii_Ggggg6     0
5,6>      .have_many_hidden_.markov_.model.s   iiii_1iii_uiii_uiiiii_1iiiii_1gggg6     0
5,6>      .have_.many_hidden_markov_.model.s   iiii_1iii_1iii_uiiiii_uiiiii_1gggg6     0
5,6>      .have_.many_hidden_.markov_model.s   iiii_1iii_1iii_uiiiii_1iiiii_Ggggg6     0
5,6>                .hidden_.markov_.model.s   iiii_iiii_iiii_1iiiii_1iiiii_1gggg6     0
\end{verbatim}
}

Finally, we assign costs (i.e., weights) to the different operations
by means of a 1-WFSM generated from the expression
\begin{eqnarray}
  A_5\tapnum{1}
  & = &
  \left(
  \aTuple{\_, 0} \cup
  \aTuple{{\tt i} , 0} \cup
  \aTuple{{\tt u} , 2} \cup
  \aTuple{{\tt g} , 1} \cup
  \aTuple{{\tt G} , 3} \cup
  \right.		\nonumber \\
  & ~ &
  \left.
  \aTuple{{\tt 1} , 0} \cup
  \aTuple{{\tt 2} , 1} \cup
  \aTuple{{\tt 3} , 1.5} \cup
  \aTuple{{\tt 4} , 2} \cup
  \aTuple{{\tt 5} , 2.5} \cup
  \aTuple{{\tt 6} , 3} \cup
  \aTuple{{\tt 7} , 3.5} \cup
  \aTuple{{\tt 8} , 4}
  \right)
\end{eqnarray}

\noindent
Here we chose the costs by intuition.
In an improved approach they could be estimated from data.

To obtain our acronym-meaning extractor,
we join $A_5\tapnum{1}$ with the previously compiled $A_4\tapnum{6}$:
\begin{equation}
  {\sf Acro}\tapnum{6} =  A_4\tapnum{6} \JOIN{6=1} A_5\tapnum{1}
\end{equation}

If we apply ${\sf Acro}\tapnum{6}$ with tapes 1 and 2 to a text chunk
and a corresponding acronym, respectively,
we obtain from tapes 5 and 6 for example

{\scriptsize
\begin{verbatim}
1,2> they_have_many_hidden_markov_models     hmms

5,6> t.hey_have_many_hidden_.markov_.model.s   G2ii_uiii_uiii_uiiiii_1iiiii_1gggg6     17
5,6> t.hey_have_.many_hidden_markov_.model.s   G2ii_uiii_1iii_uiiiii_uiiiii_1gggg6     17
5,6> t.hey_have_.many_hidden_.markov_model.s   G2ii_uiii_1iii_uiiiii_1iiiii_Ggggg6     18
5,6>      .have_many_hidden_.markov_.model.s   iiii_1iii_uiii_uiiiii_1iiiii_1gggg6     11
5,6>      .have_.many_hidden_markov_.model.s   iiii_1iii_1iii_uiiiii_uiiiii_1gggg6     11
5,6>      .have_.many_hidden_.markov_model.s   iiii_1iii_1iii_uiiiii_1iiiii_Ggggg6     12
5,6>                .hidden_.markov_.model.s   iiii_iiii_iiii_1iiiii_1iiiii_1gggg6      7
\end{verbatim}
}

\noindent
We select the analysis with the lowest cost by means of a
classical single-source best-path algorithm such as 
Dijkstra's algorithm \cite{dijsktra:1959}
or Bellman-Ford's \cite{bellman:1958,ford+fulkerson:1956}.
If our weights have been optimally chosen, we should now obtain the correct analysis.

In practice, input is read on tapes 1 and 2 and output generated from tapes 2 and 5,
as in the following examples.
All other tapes can therefore be removed, leaving us with a 3-WFSM ${\sf Acro}\tapnum{3}$.

{\scriptsize
\begin{verbatim}
1,2> they_have_many_hidden_markov_models    hmms
1,2> between_hidden_markov_models           hmms
1,2> and_weighted_finite_state_automata     wfsa
1,2> and_weighted_finite_state_automata     wfa

2,5> hmms    .hidden_.markov_.model.s
2,5> hmms    .hidden_.markov_.model.s
2,5> wfsa    .weighted_.finite_.state_.automata
2,5> wfa     .weighted_.finite_state_.automata
\end{verbatim}
}


\section{Some Results}

We tested the acronym-meaning extractor on many examples.
Finding the best analysis for one acronym took only a few milliseconds (ms)~:
For example,
3.7 ms for
``{\small\sf they have many hidden markov models}''-``{\sf\small hmms}'',
9.6 ms for
``{\small\sf they have many hidden markov models they have many hidden markov models}''-``{\sf\small hmms}'',
and 12.0 ms for
``{\small\sf they have many hidden markov models they have many hidden markov models}''-``{\sf\small hmmshmms}''.

The extractor had a size of 27 states and 64 transitions.



\end{document}